\documentclass[letterpaper, 10 pt, conference]{ieeeconf}  
\IEEEoverridecommandlockouts
\usepackage{cite}
\usepackage{amsmath,amssymb,amsfonts}
\usepackage{algorithmic}
\usepackage{graphicx}
\usepackage[hyphens,spaces,obeyspaces]{url}
\usepackage{hyperref}
\usepackage{todonotes}
\usepackage{textcomp}
\usepackage{xcolor}
\usepackage{bm}
\usepackage{threeparttable}
\usepackage{siunitx}
\usepackage{multirow}
\usepackage{subfig}
\usepackage{comment}
\usepackage{tikz}
\newcommand*\circled[1]{\tikz[baseline=(char.base)]{
            \node[shape=circle,draw,inner sep=2pt] (char) {#1};}}

\usepackage[skip=8pt,font=small]{caption}

\def\BibTeX{{\rm B\kern-.05em{\sc i\kern-.025em b}\kern-.08em
    T\kern-.1667em\lower.7ex\hbox{E}\kern-.125emX}}

\hypersetup{
	colorlinks=true,
	linkcolor=black,
	filecolor=magenta,      
	urlcolor=black,
}

\begin{document}

\title{\LARGE \bf Advantages of Multimodal versus Verbal-Only Robot-to-Human Communication with an Anthropomorphic Robotic Mock Driver
}

\author{Tim~Schreiter$^1$,
       Lucas Morillo-Mendez$^1$,
       Ravi T. Chadalavada$^1$,
       Andrey Rudenko$^2$, \\
       Erik Billing$^3$,
       Martin Magnusson$^1$,
       Kai O. Arras$^2$
       and~Achim J.~Lilienthal$^{4,1}$
\thanks{$^{1}$Centre for Applied Autonomous Sensor Systems (AASS),
	\"Orebro University, Sweden {\tt\small \{tim.schreiter, lucas.morillo, ravi.chadalavada, achim.lilienthal\}@oru.se}}
\thanks{$^{2}$Robert Bosch GmbH, Corporate Research, Stuttgart, Germany
{\tt\small andrey.rudenko@de.bosch.com}}%
\thanks{$^{3}$Interaction Lab, University of Skövde, Sweden 
{\tt\small erik.billing@his.se}}%
\thanks{$^{4}$TU Munich, Germany 
{\tt\small achim.j.lilienthal@tum.de}}%
\thanks{This work was supported by the European Union’s Horizon 2020 research and innovation program under grant agreement No. 101017274 (DARKO)}}

\maketitle

\begin{abstract}
Robots are increasingly used in shared environments with humans, making effective communication a necessity for successful human-robot interaction. In our work, we study a crucial component: active communication of robot intent. Here, we present an anthropomorphic solution where a humanoid robot communicates the intent of its host robot acting as an ``Anthropomorphic Robotic Mock Driver'' (ARMoD). We evaluate this approach in two experiments in which participants work alongside a mobile robot on various tasks, while the ARMoD communicates a need for human attention, when required, or gives instructions to collaborate on a joint task. The experiments feature two interaction styles of the ARMoD: a verbal-only mode using only speech and a multimodal mode, additionally including robotic gaze and pointing gestures to support communication and register intent in space. Our results show that the multimodal interaction style, including head movements and eye gaze as well as pointing gestures, leads to more natural fixation behavior. Participants naturally identified and fixated longer on the areas relevant for intent communication, and reacted faster to instructions in collaborative tasks. Our research further indicates that the ARMoD intent communication improves engagement and social interaction with mobile robots in workplace settings.
\end{abstract}


\section{Introduction}\label{sec:intro}

In today’s workplaces, mobile robots are becoming increasingly common, working alongside human colleagues. However, while humans use a complex set of social cues to interact with each other, mobile robots are often limited by their native design, making it difficult for them to produce legible social cues. To enable mobile robots to convey critical information about their environment and the task at hand to their human co-workers, designing efficient communication methods is paramount. Therefore, ensuring seamless and productive interactions between robots and humans requires the development of suitable methods to bridge the communication gap between them.

\begin{figure}[!t]
\centering
\includegraphics[width=8cm]{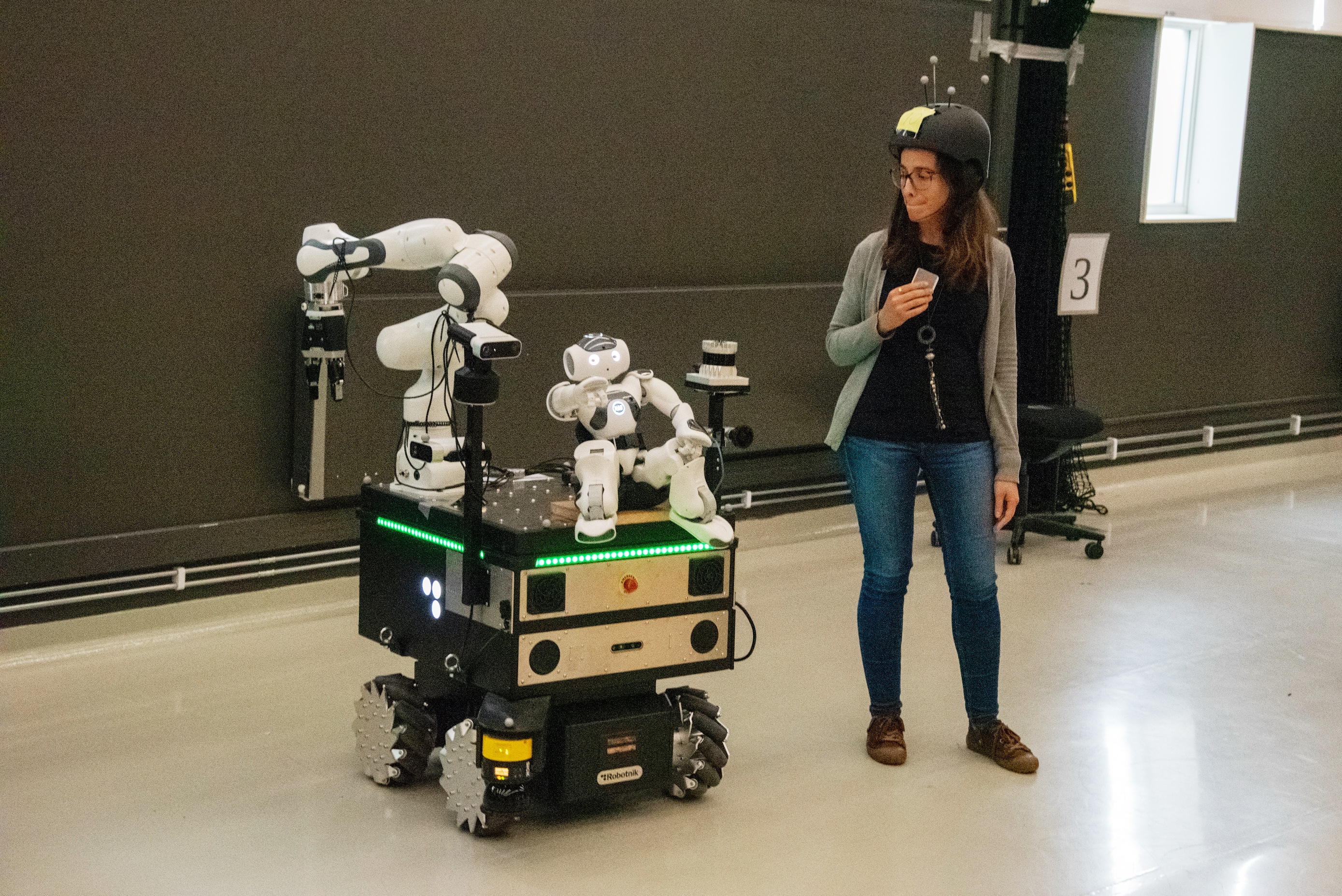}
\caption{Participant encountering a mobile robot with a NAO robot mounted on top as the ``Anthropomorphic Robotic Mock Driver'' (ARMoD). The mobile robot communicates with participants through the ARMoD.}
\label{abb:naosetup}
\vspace*{-5mm}
\end{figure}

The need for effective communication between mobile robots and humans in different work environments has led to research into various approaches, including native communication channels such as LEDs \cite{song2019designing, sanoubari2022message} and robot-attached channels such as floor projections \cite{chadalavada2020bi, linde2023}. However, these cues may not be universally understood or applicable to all robots. The need for approaches that can be validated and used across a range of mobile robots still remains open \cite{cha2018survey}. In this study, we investigate the use of an ``Anthropomorphic Robotic Mock Driver" (ARMoD) as seen in Figure~\ref{abb:naosetup} to facilitate intuitive communication between non-humanoid robots and human co-workers in workplace settings, building on the previous research in this area \cite{severinson2003social, salem2011friendly, chadalavada2020bi, schreiter2022effect}.

This paper explores the incorporation of communication channels for mobile robots using the ARMoD, without affecting their primary functionality. The ARMoD is a humanoid robot mounted on top of mobile robots for anthropomorphic intent communication. Prior research showed that adding anthropomorphic features can enhance communication with pedestrians \cite{chang2022can}. We implement the ARMoD to facilitate more natural and intuitive communication between humans and mobile robots by leveraging social cues through its anthropomorphic features. Our initial validation of the ARMoD concept concluded an increase in appearance-based trust to the robot \cite{schreiter2022effect}. In this study, we investigate the interactive capabilities of the ARMoD and examine its effects on participants’ attention by measuring their eye-gaze during the interaction in a collaborative task. To frame our experiments, we draw on the terminology of the intent communication model introduced by Pascher et al. \cite{pascher2023communicate} to categorize the robot’s conveyed intents.

In human-robot interaction (HRI), eye tracking is a powerful tool for analyzing visual attention and perception. Researchers can gain valuable insights into how people perceive and interact with their environment by recording and analyzing fixations, brief periods when the eye remains relatively stable and visual information is acquired \cite{trabulsi2021optimizing}. Previous research used eye tracking to investigate how a robot's intent communication affected human bystanders' gaze \cite{chadalavada2020bi} and participants' engagement \cite{kompatsiari2019measuring}. In our study, we use eye tracking to analyze how participants' fixations are distributed between the ARMoD and the mobile robot, and how the interaction style of the ARMoD affects participants' reaction times to cues that are relevant for collaborative tasks.

To validate the interactive capabilities of the Anthropomorphic Robotic Mock Driver as an intention communication entity for mobile robots, we design two different styles of interaction: a purely verbal one, where the intention is communicated using only the speech of the humanoid robot, and a multi-modal one, where the intention is supported by the robotic gaze and pointing gestures of the robot. These are in line with recent literature on humanoid robots \cite{salem2011friendly, kompatsiari2019measuring}. Our aim is to investigate their effect on the communication of different types of intentions to human users. The ARMoD is mounted on two different mobile robots, and interacts with human participants in either verbal-only or multimodal communication styles, depending on the experimental condition.

To investigate the impact of these two different interaction styles on the quality of human-robot interaction aided by ARMoD, we conduct two experiments in which participants work
alongside the robot on various tasks which require collaboration with the robot. In these experiments, we address the following three research questions:
\begin{enumerate}
    \item How do different interaction styles influence participants’ fixation duration on the ARMoD during an attention-grabbing greeting behavior?
    \item Does an interaction style that registers communicated intent in space lead to faster reaction times compared to a style that does not?
    \item To which extent do participants fixate on the ARMoD vs. the mobile robot during HRI and how are two different interaction styles affecting this behavior?
\end{enumerate}
Our study validates the observation of prior research by Salem et al. \cite{salem2011friendly} that a multimodal interaction style of a humanoid robot leads to participants interacting in a ``fairly natural way". Furthermore, we find additional evidence that eye contact established by a humanoid robot leads participants to longer fixate on its face, which Kompatsiari et al. \cite{kompatsiari2019measuring} correlated with increased engagement. Equipping mobile robots with an ARMoD that utilizes a multimodal interaction style to communicate with users results in faster reaction times in collaborative tasks, where the robotic gaze registration of communicated instructions enabled quicker localization of goal points and objects of interest. Our study concludes the potential for the ARMoD as a flexible HRI concept to enhance engagement and social interaction with mobile robots in workplace settings. 

\section{Related Work}

Anthropomorphic features provide rich opportunities to express social cues, making them more engaging and acceptable to users. In a study by Zlotowski et al. \cite{zlotowski2015anthropomorphism}, the authors explore the potential of anthropomorphism in human-robot interaction, highlighting the importance of developing robots that can effectively communicate and express themselves in a human-like manner. Pascher et al. \cite{pascher2023communicate} even point out that anthropomorphic elements for communicating intent share the same baselines as in human-human collaboration. The general assumption is that, in turn, they can be easily understood by users and can mostly be integrated into the actual HRI.

As such, there is great potential in exploring the idea of a proxy with anthropomorphic features for a mobile robot, to act as a natural communication partner and communicate the mobile robot's intents. The work by Severin-Eklundh et al. \cite{severinson2003social} first introduced this concept, by using an embodied interface character ``CERO'' to enhance the user experience in human-robot interactions. CERO was not a real robot, but rather a caricaturistic ``driver", however, the years after the publication have seen the development of commercially successful humanoid robots such as the NAO robot \cite{gouaillier2008nao}. The potential of using a ``proper social'' to explore the effectiveness of anthropomorphic cues in human-robot interaction, in a way CERO could not, is substantial. In this paper, we build on this idea by using a humanoid robot (NAO) to investigate the effectiveness of a multimodal interaction style, including verbal and gestural communication channels, in directing participants' attention in a task-based interaction scenario.

Recent studies have explored the potential of modern humanoid robots with anthropomorphic features in communicating intent. For instance, Salem et al. \cite{salem2011friendly} investigated two different interaction styles for a Honda humanoid robot in a domestic setting. They found that the robot was evaluated more positively when hand and arm gestures were used alongside speech. Building on this work, our study also investigates two similar interaction styles to evaluate the effectiveness of our ARMoD. However, we further extend the multimodal interaction style proposed by Salem et al. by introducing the robotic gaze as an additional modality. Recent HRI literature suggests that robotic gaze can elicit engagement \cite{kompatsiari2019measuring} and drive attention, even when their eyes are not visible to the human \cite{morillo2023}.

In summary, effective communication of intent is critical for successful human-robot interaction, and recent studies have explored different ways to achieve this. Efficient on-robot communication channels for mobile robots have been investigated, and the potential of anthropomorphic features to enhance intent communication has been highlighted. The CERO character, introduced by Severin-Eklundh et al. \cite{severinson2003social}, was an early attempt to use anthropomorphic features for this purpose, but limitations in technology at the time meant that this idea could not be fully explored. Recent advancements in technology, particularly in the development of humanoid robots such as the NAO, have enabled new possibilities for exploring the effectiveness of anthropomorphic features in HRI. Our proposed ARMoD concept builds on this idea, incorporating the robotic gaze as an additional modality for intent communication and making the interaction between humans and robots feel more natural and intuitive.

\section{Experimental Methodology and Design}\label{sec:methods}\label{sec:experiments_description}

\begin{figure}[!t]
\vspace{0.2cm}
\centering
\includegraphics[width=8cm]{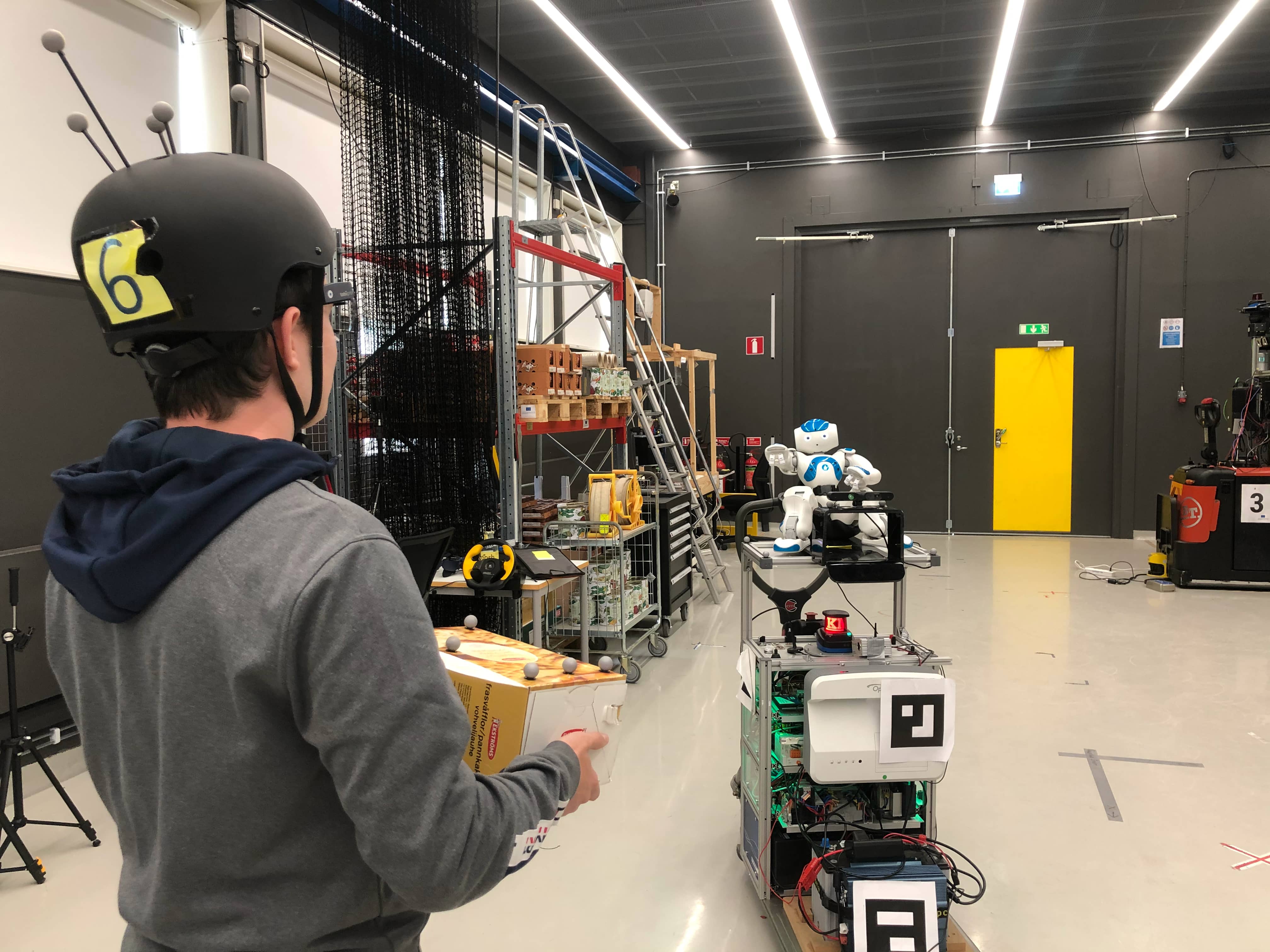}
\caption{In Experiment A, participants interact with a robotic forklift. The ARMoD instructs the participants to place an object on the forks of the mobile robot.}
\label{abb:naosetup_expA}
\end{figure}

This study explores the ability of the ``Anthropomorphic Robotic Mock Driver" (ARMoD) to communicate intentions for mobile robots in a workplace setting. We examine the impact of two interaction styles -- verbal-only and multimodal -- on conveying various intentions, including attention, motion, and instruction. In this context, attention refers to when a robot aims to catch the user’s attention for a subsequent movement activity. The ARMoD is mounted on a different mobile robot in each experiment and interacts with human participants using one of the two interaction styles based on the experimental condition. Our ARMoD, a NAO robot, is fixed to a seat for consistent positioning. This section provides a detailed description of our experimental design and methodology.

This paper presents results of two experiments to validate the ARMoD concept and answer the research questions. The initial Experiment A investigates the interaction styles of the ARMoD in one-on-one interactions in a narrow corridor. Intriguing fixation patterns are observed, such as longer fixation on the face of a humanoid robot when eye contact was established, and faster reaction times in collaborative tasks when using a multimodal interaction style with an ARMoD. The consequent Experiment B is designed to confirm these findings using a different mobile robot in repeated interactions in a more open workplace setting.

In both experiments, participants act as coworkers with the mobile robot and work alongside it on various tasks. When the robot encounters a situation in which it requires assistance to complete its task, the ARMoD communicates the need for the human's attention, with the goal of initiating an interaction. Once the interaction starts, the human becomes the collaborator in a joint task with the robot. Participants are instructed to collaborate with the robot if it is requested by the ARMoD. In both experiments, the ARMoD communicates instructional and motion intent to coordinate the fulfillment of the collaborative task with the human.

The experiments take place under two different conditions, each modulating the interaction style of the ARMoD. In the verbal-only condition, the ARMoD communicates solely verbally with the participants. In the multimodal condition, we combine verbal communication with gaze cues and pointing gestures from the NAO robot to register communicated intent in space if necessary. This multimodal interaction style builds on the one proposed by Salem et al. \cite{salem2011friendly}.

Experiment A and Experiment B differ primarily in the mobile robots used, the nature of the collaborative task, and the design of the workspace. In Experiment A, participants transport an aluminum tin can (diameter 160~mm, filled with 750 ml canned vegetables) to a table and then collaborate with a robotic forklift, which must transport a box (see Figure~\ref{abb:naosetup_expA} and Figure \ref{abb:expsetup}) to the other side of a corridor. The ARMoD instructs the humans to place the box on the forklift's forks and, once the box is loaded, guides the human's path to avoid a collision by using its voice to say ``Pass on my left'' and pointing to its left in the multimodal interaction style. In contrast, in Experiment B, participants interact with a smaller, more agile mobile robot with different physical appearance and driving characteristics, see Figure \ref{fig:evalFront}. This mobile robot, equipped with a robotic arm in its resting position, navigates in a $10\times9$ meter open workplace setting and requires the assistance of a human at a specific goal point. The ARMoD communicates the robot's next goal point and instructs the human to accompany it.

In our experiments, we use Tobii eye tracking glasses (versions 2 and 3) to capture the participants' gaze behavior during the interaction with the robots. The data obtained from the Tobii glasses requires post-processing for suitable data analysis, as we describe in Section \ref{subsec:Posthoc}. Otherwise, the results are susceptible to misinterpretation. We deploy the standard Tobii IVT attention gaze filter with a classification threshold of $100^\circ/$s. For the evaluation, we use the software "TobiiProLab"\footnote{\url{https://www.tobiipro.com/product-listing/tobii-pro-lab/}}. We describe the preparation of the eye tracking data in Section \ref{subsec:Posthoc} and its analysis in Sections \ref{sec:result} and \ref{sec:discuss}.

In addition to the eye gaze trackers, in both experiments we measure subjective ratings and perception of the robot in questionnaires.
For experiment A we deploy the same trust scale for ``Trust in Industrial Human-robot Collaboration'' by Charalambous et al. \cite{charalambous2016development} as for our prior work \cite{schreiter2022effect}, to assess how an interaction is affecting the subjective user ratings. In Experiment B, we add Bartneck’s ``Godspeed questionnaire" \cite{bartneck2009measurement} to gain a more comprehensive understanding of participants’ perception of the robot system and to check for potential differences in interaction styles. 

\subsection{Experiment A: Request of human assistance}

\begin{figure}[t]
\vspace{0.2cm}
\centering
\includegraphics[width=8cm]{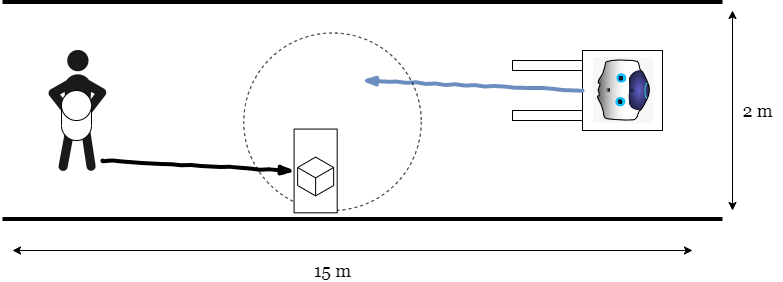}
\caption{Experimental setup for \textbf{Experiment A}, in which a human participant interacts with an Anthropomorphic Robotic Mock Driver (ARMoD) seated on a mobile robotic forklift. The participant begins at one end of a corridor, the forklift and ARMoD at the opposite end. The experiment involves the task to transport a tin can and later collaborate with the robot to place a box according to instructions on the forklift.}
\label{abb:expsetup}
\end{figure}

In Experiment A, we explore the two interaction styles of the ARMoD giving simple instructions. To counterbalance learning effects, each participant takes part in both conditions in random order. One interaction style is verbal-only, while the other is multimodal and includes pointing gestures, robotic gaze, and eye contact with participants. The interactions take place in a 15~m long and 2~m wide corridor. Participants approach a table to pick up a box and correctly place it on a marked area on the robot’s forks. The ARMoD then instructs participants to disengage. The interaction is initialized when the distance between the participant and the forklift is less than or equal to five meters, based on the social distance model by Hall \cite{hall_article}. Prior to the experiment, a human instructor explains how to place objects on the robotic forklift’s forks as participants are not expected to have prior experience with forklifts. Figure \ref{abb:expsetup} shows the experimental setup.

\begin{figure}[!t]
\vspace{0.2cm}
\centering
\includegraphics[width=7cm]{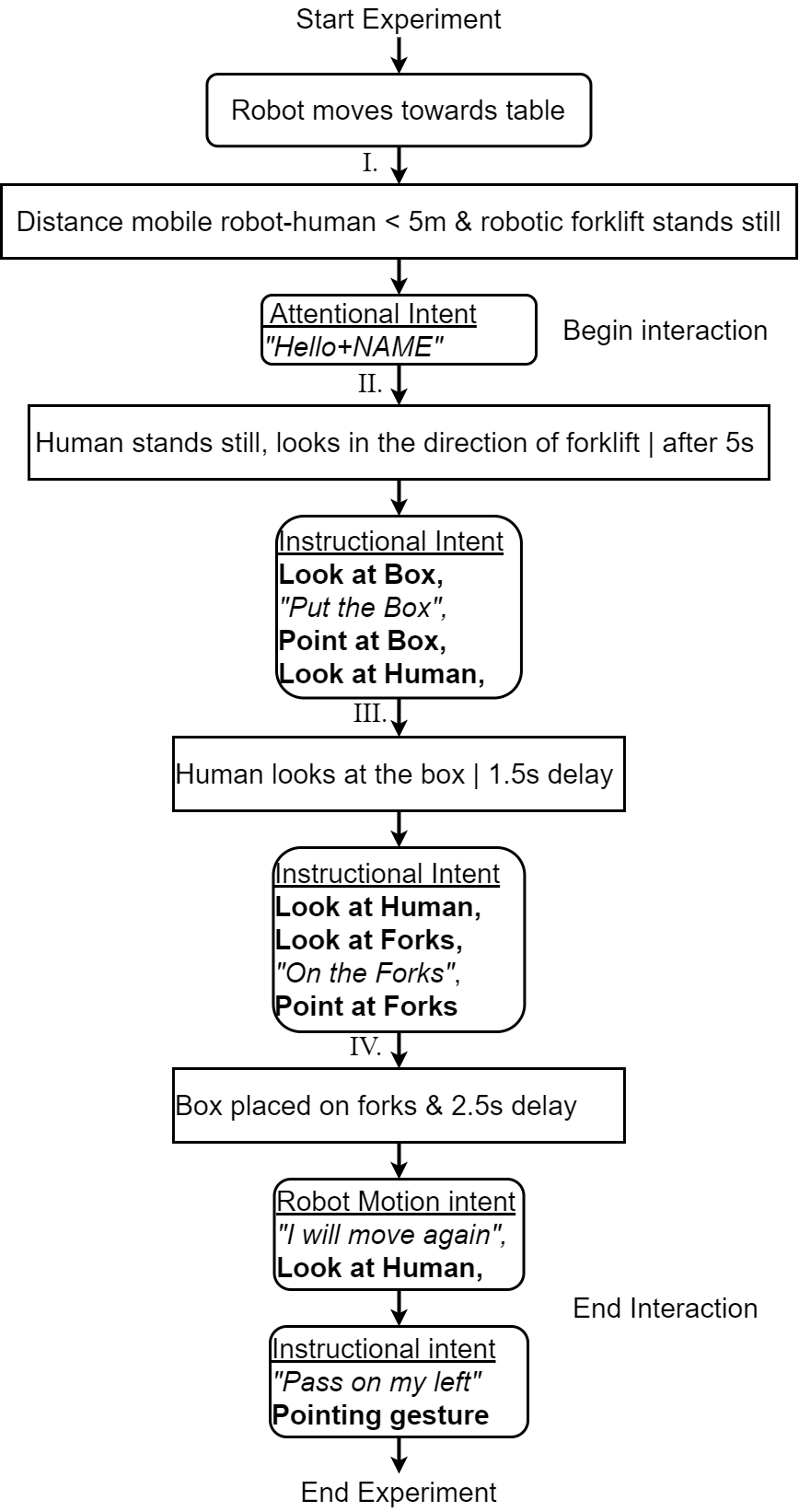}
\caption{Flow chart illustrating the programmed behavior of the ARMoD during \textbf{Experiment A} in a hallway encounter. The sequence of events during each step of the interaction is shown from top to bottom. Dialogue spoken by the ARMoD is indicated by \textit{italicized text in quotes}, while \textbf{bold text} indicates movements that are only present in the multimodal interaction style condition.}
\label{abb:paplandscape}
\end{figure}

The ARMoD deploys various gazes and gestures during the interaction with participants. When the ARMoD's distance to the participant is less than or equal to five meters, the ARMoD starts giving instructions. In the multimodal interaction style, the ARMoD performs referential gestures and gazes while speaking, making eye contact with the participants, and tracing them using head movements. The spoken instruction ``Pass on my left'' is accompanied by an optional referential gesture. In the verbal-only interaction style, the ARMoD only gives spoken instructions while looking in the driving direction. The program sequence plan, shown in Figure \ref{abb:paplandscape}, details the sequence of actions and behaviors of the ARMoD during the interaction with participants in Experiment A. Interactions ranged from 74~s to 104~s with a median duration of 89~s in with the verbal-only and 96~s with the multimodal interaction style.

\subsection{Experiment B: Mediating joint navigation}\label{subsec:interact}

\begin{figure}[!t]
\vspace{0.2cm}
\centering
\includegraphics[width=5cm]{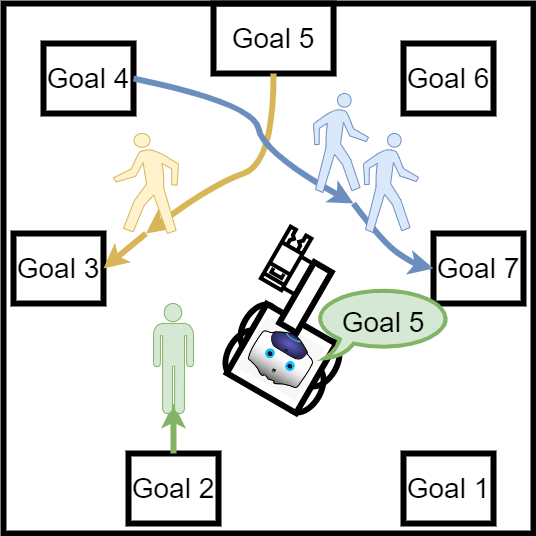}
\caption{Experimental setup for \textbf{Experiment B}, which investigates the interaction between multiple participants and robots in a shared workplace setting. Participants navigate between designated goal points by drawing cards, as described in \cite{thor2020, schreiter2022magni}. Two special cards instruct participants using the phrase ``Go to the robot'' to look for the robot, approach and interact with it. The study aims to examine participants' behavior and perceptions during these interactions in a dynamic, realistic environment.}
\label{fig:magni_expsetup}
\end{figure}

Experiment B verifies Experiment A’s findings by testing the interaction styles with a different mobile robot and repeated interactions. It evaluates the difference between multimodal and verbal-only styles for collaborative tasks and compares user ratings and perceptions. Participants navigated freely with the robot in an open room with seven goal points (see Figure~\ref{fig:magni_expsetup}). Participants drew cards from decks at designated goal points which indicated their next navigation goal. Each deck had a varying number of cards, with goal points \circled{1} and \circled{7} having 15 cards each, goal point \circled{3} having 12 cards, and goal points \circled{4}, \circled{5}, and \circled{6} having 9 cards each. Two special cards instructed participants to look for the robot in the room and interact with it.

Upon encounter, the ARMoD initiated the interaction in either a multimodal or verbal-only style. An experimenter monitored the scene and adjusted the ARMoD's behavior by entering the next goal point for the mobile robot. This was communicated to participants through the ARMoD. If too many participants were at a goal point, the experimenter interrupted the mobile robot’s autonomous navigation shortly before reaching it. If interrupted prematurely, the mobile robot would tell the participant to abort the interaction and continue drawing cards. The mobile robot would navigate alone to the goal point once it was less crowded.

In Experiment B, we explore the two interaction styles (multimodal and verbal-only) of the ARMoD giving simple instructions. In the multimodal style, the ARMoD greeted the participant while establishing eye contact, communicated attention intent and used head and pointing gestures to instruct the participant to go to the next goal point and draw a card. At the goal point, the ARMoD again used head and pointing gestures to instruct the participant to go to the goal on the card. In the verbal-only style, the ARMoD only greeted the participant and provided final instructions at the goal point, without eye contact, robotic gaze, or pointing gestures. Depending on the interaction style and the distance between goals, interactions lasted around 30-40 seconds, with a median duration of 37 seconds for the multimodal style and 32 seconds for the verbal-only style.

\subsection{Eye Tracker recordings}\label{subsec:Posthoc}

We generate heatmaps from the eye tracking data to analyze how the different interaction styles influence participants' attention patterns and reaction times to ARMoD's instructions. To obtain these heatmaps, we label important events within the recordings captured by the Tobii Pro Glasses camera and use Tobii Pro Lab's assisted mapping tool to map the user's gazes from the eye-tracker global camera onto 2D images. We also use the software's AOI (area of interest) annotation tool to define regions of interest within the snapshots, allowing us to analyze fixation count and duration on certain robot parts. Finally, we generate heatmaps for the count of fixations of participants on the snapshots (see Figure \ref{fig:evalFront}). This process enables us to analyze the influence of interaction styles on participants' eye gaze and reaction times.

\subsection{Participants}\label{subsec:samples}

We recruited 25 participants for Experiment A and 9 for Experiment B. Participants’ ages ranged from 18 to 56 years (M = 28.7, SD = 7.88) in Experiment A and from 23 to 38 years (M = 30.2, SD = 4.73) in Experiment B. All participants are fluent in English and identify as female (14/25; 4/9), male (10/25; 5/9) or non-binary (1/25; 0/9). In Experiment B, participants interact twice with each interaction style in four four-minute long sessions in randomized order. In Experiment A, participants interact once with each interaction style in two-minute long sessions.

\section{Results}\label{sec:result}

\begin{figure}[!t]
\vspace{0.2cm}
     \centering
     \subfloat[][Heatmap condition verbal-only interaction style]{\includegraphics[width=4cm]{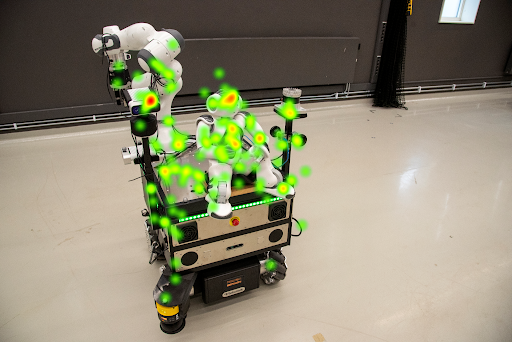}\label{fig:MagniACount}}\hfill
     \subfloat[][Heatmap condition multimodal interaction style]{\includegraphics[width=4cm]{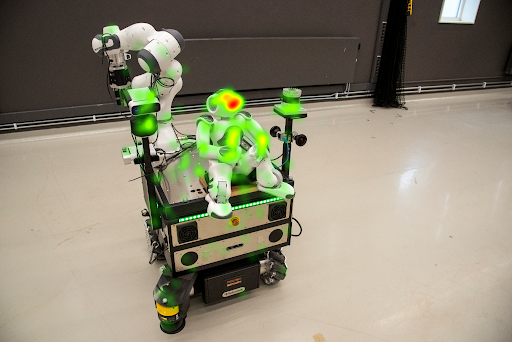}\label{fig:MagniBCount}}
     \caption{Heatmaps showing participant gaze distribution on the robot platform in two conditions. In the \textbf{verbal-only condition (left)}, fixations are spread widely across the robot and its sensory equipment, with multiple red blobs on the ARMoD's body and one on the RGBD camera. In the \textbf{multimodal condition (right)}, participants focus more strongly on the ARMoD, as indicated by the single red blob on the robot's face. Red blobs indicate centers of high fixation counts in both heatmaps.}
     \label{fig:evalFront}
\end{figure}

We present the results of qualitative questionnaires and quantitative eye tracking measurements. The questionnaires provide limited insights due to the small and heterogeneous sample size. Therefore, we primarily rely on the eye tracking measurements to address our research questions.  

\subsubsection{Questionnaires}

We gathered subjective user ratings in the system using Charalambous' questionnaire \cite{charalambous2016development} for ``Trust in Industrial Human-robot Collaboration" in both experiments. In Experiment A, we tested for significant differences in subjective user ratings between the two proactive interaction styles with different modalities and the data for an interaction with no modalities from our prior work \cite{schreiter2022effect} using a one-way ANOVA. The median scores were 42 for the interaction with no modalities and 43 for both verbal-only and multimodal interactions. This may indicate a slight improvement in subjective trust using either interaction style. However, no significant difference between the groups was found in the statistical test (F-statistic = 0.22, p = 0.80).

In Experiment B, we added Bartneck's Godspeed questionnaire to evaluate participants' subjective perceptions of the ARMoD's interaction styles. We used a Mann-Whitney U tests to compare sub-scales between verbal-only and multimodal interaction styles. The analysis shows small, non-significant differences for some constructs in the questionnaire. We use Shapiro-Wilk tests to confirm that all data was not normally distributed before performing the tests. Results show no significant difference between the groups in any of the subscales (all p-values $>$ 0.05). The median scores are 10 for both conditions in the Anthropomorphism subscale, 13 for verbal-only and 16 for multimodal in the Animacy subscale, 18 for both conditions in the Likeability subscale, 14 for verbal-only, and 15 for multimodal in the Intelligence subscale. Each subscale is rated on a scale from 1 to 25, with higher scores indicating more positive levels of the attribute being measured. For the Safety subscale (1 to 15) they are  10 for the verbal-only and 11 for the multimodal interaction style. 

\subsubsection{Gaze Behavior of Participants during the interactions}\label{subsec:eyt}

We found that participants fixated on the robots differently between the verbal-only and multimodal interaction styles in both experiments. Figure \ref{fig:evalFront} shows sample heatmaps generated from the gaze data of participants in experiment B. The heatmap on the left (Figure \ref{fig:MagniACount}) for the verbal-only interaction style shows scattered fixation counts across the robots, while the heatmap on the right (Figure \ref{fig:MagniBCount}) for the multimodal interaction style shows a large center of high fixation counts around the head of the robot. Similarly, in experiment A, the heatmaps show a clear focus on the head of the ARMoD for the multimodal interaction style. With the absolute fixation count per heatmap, we calculate how much percent of these fixations land in certain regions of interest. With the respective median durations of interactions, we calculate the fixation frequencies as 1.62~Hz and 1.67~Hz for the multimodal and 2.83~Hz and 2.5~Hz for the verbal-only interaction style in Experiments A and B.

\begin{figure}[t]
\vspace{0.2cm}
\centering
\includegraphics[width=8cm]{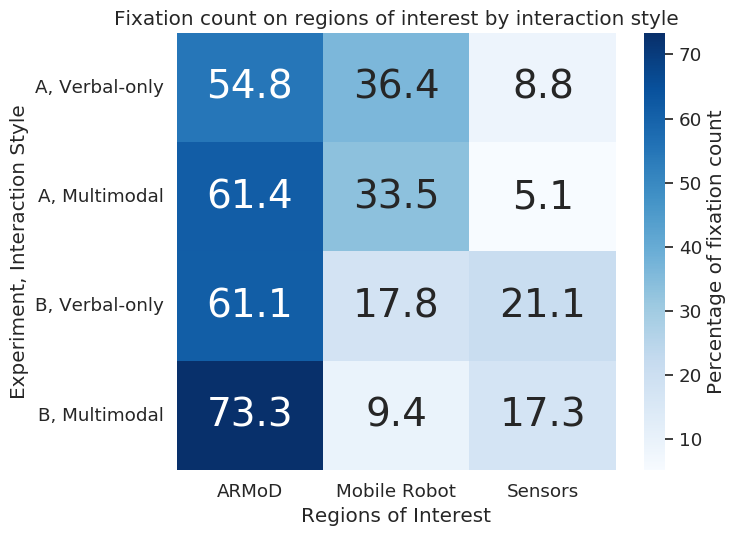}
\caption{Matrix comparing the percentages of fixation counts on regions of interest for verbal-only and multimodal interaction styles. Fixations on the background or other parts of the scene that receive a very little amount of fixations are excluded from the analysis to focus on how participants fixate on the robots during the interaction. In the multimodal interaction style, the ARMoD receives more fixations, suggesting that participants interact with it in a ``fairly natural way'' (as per Salem et al.~\cite{salem2011friendly}).}
\label{abb:fixviz}
\end{figure}

Figure \ref{abb:fixviz} shows the percentage of the total fixation count for each region of the analyzed heatmaps in the experiments. Verbal-only interaction saw a higher fixation count on the platform and sensors, while multimodal interaction saw a higher percentage of fixations on the ARMoD. T-tests found significant differences between verbal-only and multimodal interaction styles for both ARMoD ($p = 0.01$) and Mobile Robot AOI counts ($p = 0.02$), with small and medium effect sizes (Cohen's $d$: 0.29 and 0.49). These results suggest that the presence of visual and gestural cues in the multimodal interaction style shifts participants' fixations towards the ARMoD as the entity communicating intent. This finding is consistent with the heatmap analysis and further supports the effectiveness of the multimodal interaction style in directing participants' attention toward the communication interface.
 
We also analyzed the duration of fixations on the ARMoD based on its interaction style. The duration of all fixations during the interactions with the ARMoD was extracted for each condition in the two experiments. Independent t-tests were then performed for each condition to test for statistical significance. Participants underwent the conditions in a randomized order to counterbalance learning effects. During Experiment A, participants fixated slightly longer on the ARMoD (M $= 232$ ms, SD $= 159$ ms) in the multimodal interaction style than in the verbal-only interaction style (M $= 226$ ms, SD $= 153$), although this difference was not statistically significant ($t = 0.77$, $p = 0.44$). However, in Experiment B, we found a statistically significant difference ($t=-3.38$, $p=0.01$, Cohen's $d=0.34$) between the mean fixation duration of verbal-only and multimodal interaction styles of the ARMoD. Participants fixated significantly longer on the robot during the multimodal interaction style (M $=278$ ms, SD $=192$) than during the verbal-only interaction style (M $=212$ ms, SD $=136$).

\begin{figure}[t]
\centering
\includegraphics[width=7.5cm]{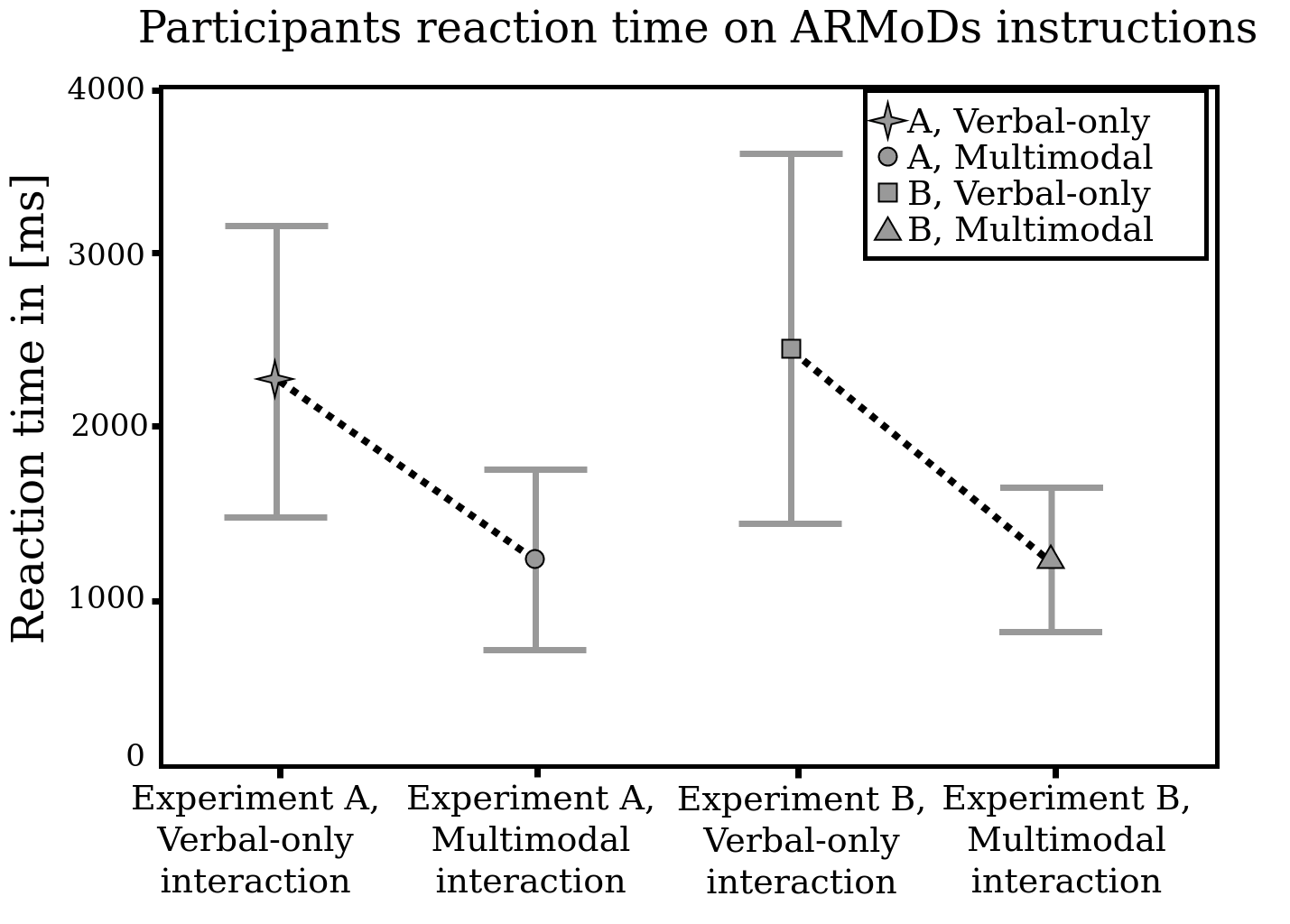}
\caption{Lineplot to compare the reaction time of participants between instruction of the ARMoD and the first fixation on the target. Error bars show standard deviation. \textbf{Left:} Experiment A, ARMoD gave instructions to place a box. \textbf{Right:} Experiment B, ARMoD gave instructions regarding the next common goal point.}
\label{abb:CompareReactionTimes}
\end{figure}

We analyzed the participants’ time to first fixation on a point or object of interest after a world-centered instructional intent communicated by the ARMoD in both experiments. Using two event markers, we measured the time between the instruction and the first fixation on the target point or object. We tested the data for normality using the Shapiro-Wilk test and deployed a Mann-Whitney U test as the Shapiro-Wilk tests did not indicate normality. Our analysis found significantly shorter times to first fixation (reaction times) for the multimodal interaction style compared to verbal-only in both experiments. Experiment A showed a decrease from M $=2317$ ms, SD $=853$ (verbal-only) to M $=1237$ ms, SD $=524$ (multimodal), a difference of 1080 ms ($U=6$, $p=0.03$, Cohen’s $d=0.39$). Experiment B showed a decrease from M $=2505$, SD $=1095$ (verbal-only) to M $=1232$, SD $=436$ (multimodal), a difference of 1273 ms ($U=9$, $p=0.03$, Cohen’s $d=0.41$). Figure \ref{abb:CompareReactionTimes} illustrates the decrease in reaction time. These results suggest that multimodal interaction styles facilitate faster and more efficient communication of intent compared to verbal-only. 

\section{Discussion}\label{sec:discuss}

\subsection{Subjective user ratings}

In our previous work, we conducted a study on hallway encounters with a mobile robot equipped with an ARMoD, which revealed higher levels of trust from participants \cite{schreiter2022effect}. In this study, we examined subjective user ratings for verbal-only and multimodal interaction styles during collaborative tasks. We found no statistically significant difference in subjective user ratings between the two interaction styles. However, in terms of animacy, likability, intelligence, and safety, the multimodal style showed slightly higher median ratings. These findings are in support of previous research by Salem et al. \cite{salem2011friendly}, who observed that participants had more positive perceptions and evaluations of robots with multimodal interaction styles. Previous user studies evaluating the text-to-speech, appearance, and performance of the NAO robot have shown that users desire more natural speech and gesture capabilities \cite{amirova21}. Therefore, future research could investigate how the subjective evaluations of the users would vary with a more sophisticated robot, such as the iCub robot used by Kompatsiari et al. \cite{kompatsiari2019measuring}.

\subsection{Gaze Behavior of Participants during Interactions}

The results of both experiments support the idea that eye contact can ``freeze attentional focus on the robot’s face'' \cite{kompatsiari2019measuring}, suggesting that incorporating head movements and robotic gaze cues into the ARMoD’s interaction style could enhance its ability to engage users. This finding addresses our first research question: ``How do different interaction styles influence participants’ fixation duration on the ARMoD during an attention-grabbing greeting behavior?". A multimodal interaction style, which includes gaze cues and eye contact for the ARMoD, may be more effective in capturing and holding participants’ attention than verbal-only interactions. This is supported by previous research \cite{kompatsiari2019measuring} that suggests that eye contact is a crucial factor in facilitating engagement and social interaction with robots. Therefore, the combination of a verbal greeting and establishing eye contact via head movements might be sufficient for the necessary attention-grabbing behaviors described by Pascher et al. \cite{pascher2023communicate} to precede the delivery of motion and instructional intents.

The effect of ARMoD, registering instructional intent in space, on participants' reaction times was examined according to our second research question: ``Does an interaction style that registers communicated intent in space lead to faster reaction times compared to a style that does not?". Two styles of interaction used by the ARMoD were compared: a verbal-only style, and a multimodal style in which the robot used head movements and pointing gestures to register intent. Pascher et al. \cite{pascher2023communicate} argue that unregistered intent requires additional mental steps to establish a spatial link, potentially slowing reaction times. Specifically, the use of head movements and robotic gaze in the multimodal interaction style appears to play an important role in this effect. Participants took 0.8 – 1~s less to fixate on an ARMoD-referenced target when these cues were used. However, the relative contributions of head movements and pointing gestures to this effect cannot be determined from this study and require further investigation. This finding is particularly relevant to industrial HRI contexts, where fast and effective communication is critical for productivity and safety.

The analysis of the heatmaps generated from participants' gaze data revealed that the multimodal interaction style was more effective at capturing and directing participants' attention than the verbal-only interaction style. This finding is in line with our third research question: ``To which extent do participants fixate on the ARMoD and the mobile robot during HRI and how are two different interaction styles affecting this behavior?". The heatmaps of Figure \ref{fig:evalFront} show that a majority of fixations were on the ARMoD's face, particularly in the multimodal interaction style. These results are in line with previous research by Gullberg and Holmqvist \cite{gullberg1999keeping}, which suggests that participants tend to fixate on a speaker's face rather than their gestures during interactions. Our findings suggest that the multimodal interaction style, with its use of head movements and pointing gestures, can effectively direct participants' attention to important spatial cues while maintaining a natural interaction style. Overall, our results highlight the potential of an ARMoD deploying a multimodal interaction style in enhancing human-robot interactions by improving attentional focus and facilitating natural communication.

\section{Conclusion and Future Work}\label{sec:concl}

Our study investigates the effectiveness of the Anthropomorphic Robotic Mock Driver (ARMoD) in providing additional communication channels for mobile robots and its impact on spatial human-robot interaction and perception of the robot by humans. We address three research questions on the influence of multimodal interaction styles on participants’ fixation duration, reaction times, and count of fixations on the ARMoD and mobile robot. We find that using an ARMoD in a multimodal interaction style leads to fewer fixations on the mobile robot and more and longer fixations on the ARMoD’s face, and shorter reaction times to communicated instructions. These results suggest that an ARMoD can effectively direct attention, and enhance communication in industrial human-robot interaction. Our study contributes to the field of human-robot interaction by providing insights into how to design optimal communication pathways for mobile robots.

This study's limitations suggest potential avenues for future research. A small sample size of participants and a bias toward having academic background may have impacted the ability to detect statistically significant differences in some of the results. Conducting experiments with a larger and more diverse sample of participants could improve the robustness of the findings. Additionally, the experiments were conducted in a controlled laboratory setting, which may limit the generalizability of the findings to real-world industrial environments. Future research could benefit from conducting experiments in real-world industrial environments to improve the generalizability of the findings. Finally, the study only used one type of ARMoD (NAO robot), which may limit the generalizability of the findings to other types of robots acting as ARMoDs. Testing different types of robots as ARMoDs could improve the generalizability and applicability of the findings. In our experiments, participants were only exposed to the robot for a brief period. In real-world applications, however, users will interact repeatedly with different types of ARMoDs, for longer periods and on a daily basis. To better understand how repeated exposure to ARMoDs and variations in their design impact user perception, future work should include long(er)-term studies with participants who repeatedly interact with different types of ARMoDs over a prolonged time. 

Our research demonstrates the potential benefits of an ARMoD for improving human-robot communication in the workplace. Future research could explore the synergies between the ARMoD and its mobile base, investigate the use of color coding and LED flashing in the ARMoD's eyes to communicate internal states and extend the use of the ARMoD to other applications beyond industrial contexts. In addition, our forthcoming data set \cite{schreiter2022magni} will provide a valuable resource for researchers investigating the prediction of human movement in the presence of an ARMoD on a mobile robot in a workplace environment. The ARMoD concept has the potential to improve the interaction between humans and robots in a wide range of domains. Further exploration of ARMoD applications in other industries and settings, such as healthcare, education, or entertainment, could lead to new and innovative ways to improve human-robot collaboration and productivity.  

\section{Acknowledgement}

We are grateful for the support of Chittaranjan Swaminathan, Janik Kaden and Timm Linder in setting up the software, Per Sporrong for technical assistance in configuring the hardware, Per Lindström for creating the mock driver seat used in this study. Their contributions were invaluable to the success of this research.

\bibliographystyle{IEEEtran}
\bibliography{IEEEabrv,references}

\end{document}